%% file: paper.tex
\documentclass{article}
\usepackage{xcolor}
\usepackage{comment}

\usepackage[preprint]{neurips_2019}

\usepackage{algorithm}
\usepackage{algorithmic}
\usepackage[utf8]{inputenc} 
\usepackage[T1]{fontenc}    
\usepackage{hyperref}       
\usepackage{url}            
\usepackage{booktabs}       
\usepackage{amsfonts}       
\usepackage{nicefrac}       
\usepackage{microtype}      
\usepackage{amsmath,amsfonts, amssymb, natbib, makecell}
\usepackage{graphicx}
\usepackage{caption}
\usepackage{subcaption}
\usepackage{relsize}

\title{Deep Reasoning Networks:  \\ Thinking Fast and Slow}

\author{%
    Di Chen\\
    Cornell University\\
    Ithaca, NY 14853 \\
    \texttt{di@cs.cornell.edu} 
   \And 
  Yiwei Bai \\
  Cornell University\\
  Ithaca, NY 14853 \\
  \texttt{bywbilly@cs.cornell.edu} \\
  \And
  Wenting Zhao \\
  Cornell University\\
  Ithaca, NY 14853 \\
  \texttt{wzhao@cs.cornell.edu} \\
    \And 
    Sebastian Ament \\
    Cornell University\\
    Ithaca, NY 14853 \\
    \texttt{ament@cs.cornell.edu} \\
    \And
    John M. Gregoire \\
    California Institute of Technology \\
    Pasadena CA 91125 \\
    gregoire@caltech.edu \\
    \And
    Carla P. Gomes \\
    Cornell University\\
    Ithaca, NY 14853 \\
    \texttt{gomes@cs.cornell.edu} \\
}

\begin{document}

\maketitle

\input{abstract}
\input{introduction}
\input{related}
\input{model}
\input{experiments}
\input{conclusion}
\subsubsection*{Acknowledgments}
This work was supported by  NSF awards CCF-1522054 and CNS-0832782 (Expeditions), CNS-1059284 (Infrastructure), and IIS-1344201 (INSPIRE); ARO award W911-NF-14-1-0498 and W911NF-17-1-0187 for the computational experiments (DURIP); AFOSR Multidisciplinary University Research Initiatives (MURI) Program FA9550-18-1-0136; an award from Toyota Research Institute; and US DOE Award No. DE-SC0004993.


\clearpage
\small
\bibliography{dichen}
\bibliographystyle{unsrtnat}
\end{document}

%% file: abstract.tex
\begin{abstract}
We introduce Deep Reasoning Networks (DRNets), an end-to-end framework that combines deep learning with reasoning for solving complex tasks, typically in an unsupervised or weakly-supervised setting. 
DRNets exploit problem structure and prior knowledge by tightly combining logic and constraint reasoning with stochastic-gradient-based neural network optimization.
We illustrate the power of DRNets on de-mixing overlapping hand-written Sudokus (Multi-MNIST-Sudoku) and on a substantially more complex task in scientific discovery that concerns inferring crystal structures of materials from X-ray diffraction data under thermodynamic rules (Crystal-Structure-Phase-Mapping). 
At a high level, DRNets encode a structured latent space of the input data, which is constrained to adhere to prior knowledge by a reasoning module.
The structured latent encoding is used by a generative decoder to generate the targeted output. Finally, an overall objective combines responses from the generative decoder (thinking fast) and the reasoning module (thinking slow), which is optimized using constraint-aware stochastic gradient descent. 
We show how to encode different tasks as DRNets and demonstrate DRNets' effectiveness with detailed experiments: DRNets significantly outperform the state of the art and experts' capabilities on Crystal-Structure-Phase-Mapping, recovering more precise and physically meaningful crystal structures. 
On Multi-MNIST-Sudoku, DRNets perfectly recovered the mixed Sudokus' digits, with 100\% digit accuracy, outperforming the supervised state-of-the-art MNIST de-mixing models. 
Finally, as a proof of concept, we also show how DRNets can solve standard combinatorial problems -- 9-by-9 Sudoku puzzles and Boolean satisfiability problems (SAT), outperforming other specialized deep learning models.
DRNets are general and can be adapted and expanded to tackle other tasks.
\end{abstract}

%% file: introduction.tex
\section{Introduction}

Human thought consists of  two different types  of processes 
\citep{kahneman2011thinking}:
\textit{System 1}, a fast, implicit (automatic), 
unconscious process, and \textit{System 2}, a slow, explicit (controlled), conscious process.
%
Humans use \textit{System 1} most of the time. \textit{System 1} is
fast,
effortless, and provides a type of near-automatic pattern recognition.
In contrast,  \textit{System 2} is slow, rational, requiring more careful thinking, and is used to solve more complex reasoning problems.
%
%

Deep learning has achieved tremendous success in 
areas such as vision, speech recognition, language translation, and autonomous driving.
Nevertheless, 
certain limitations of deep learning are 
generally recognized, in particular, limitations due to the fact that deep learning approaches heavily depend on the availability of large amounts of labeled data.
In fact, the current state of the art of 
deep learning 
has been compared to \textit{System 1}, 
 i.e., performing  pattern recognition or heuristic evaluation.
So, when it comes to complex problems that involve reasoning 
(\textit{System 2}),
such as playing Go or crystal structure phase mapping, pure machine learning approaches have to be complemented  with reasoning  algorithms, such as Monte Carlo tree search
\citep{anthony2017thinking, silver2016mastering, silver2018general}, 
or mixed-integer programming \citep{ermon2015pattern}.
Such reasoning approaches are in general outsourced using external modules, which 
is not always possible and may result in inferior performance due to the coordination barrier between neural networks (\textit{System 1}) and the outsourced reasoning module (\textit{System 2}), which is often non-differentiable.
Therefore, an efficient scheme is needed to integrate the two systems {\it in a general and seamless way. }

\begin{figure}[t]
\centering
\includegraphics[width=13.8cm]{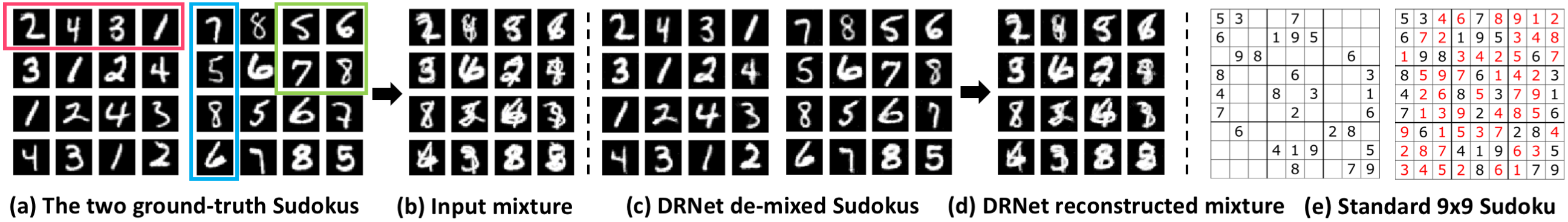}
\caption{\textbf{(a)} {Two 4x4 Sudokus: The cells in each row, column, and any of the four 2x2 boxes involving the corner cells have non-repeating digits. \textbf{(b)} Two overlapping Sudokus, with a mixture of two digits in each cell: one from 1 to 4 and the other from 5 to 8. In \textbf{Multi-MNIST-Sudoku}, the digits of two overlapping hand written Sudokus (b) have to be de-mixed 
(as done by DRNet in \textbf{(c)}).
\textbf{(d)} The reconstructed overlapping hand written Sudokus from DRNet.
\textbf{(e)} A standard \textbf{9-by-9 Sudoku puzzle:} a partially filled Soduku has to be completed as a valid Sudoku.}
}
\label{fig:sudoku_examples}
\vspace{-10pt}
\end{figure}
\begin{figure}[t]
\centering
\includegraphics[width=13.0cm]{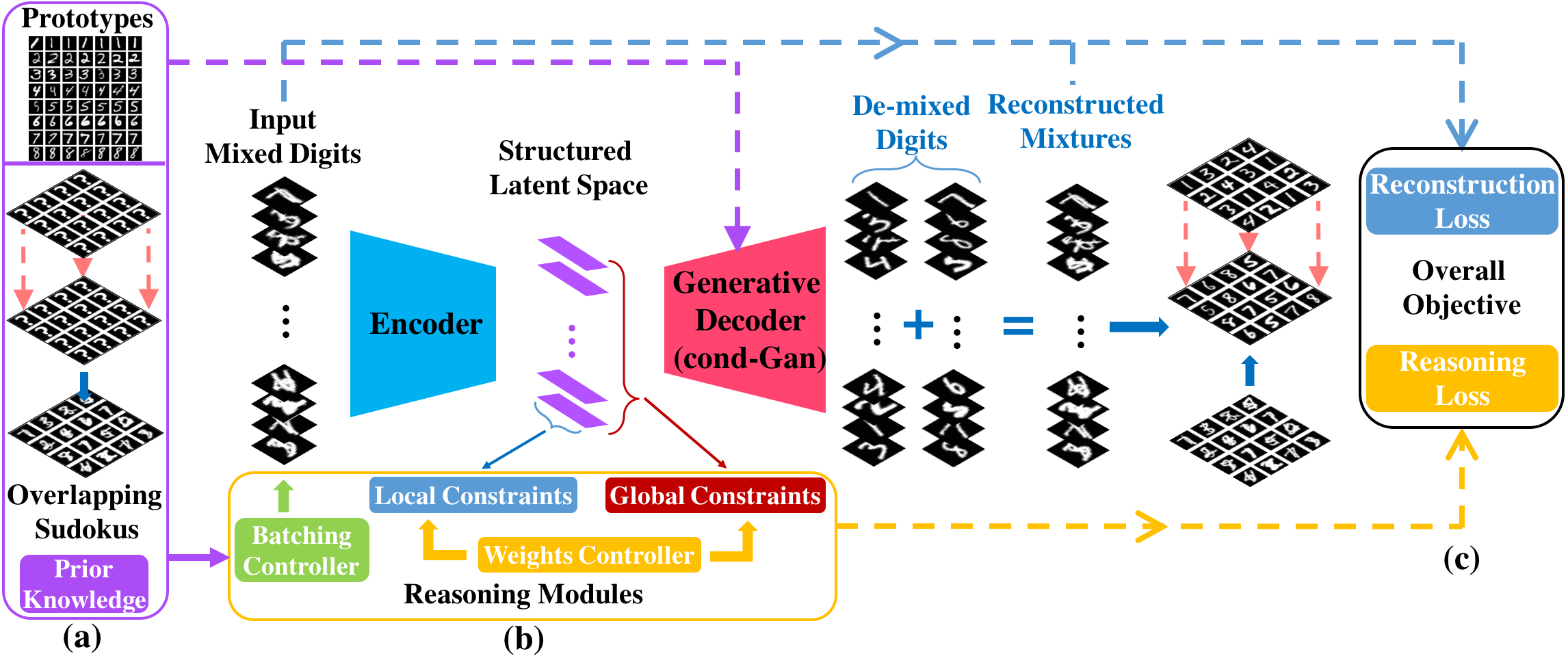}
\caption{
Deep Reasoning Networks (DRNets) perform end-to-end deep reasoning by encoding a latent space of the input data that captures prior knowledge constraints and is used by a generative decoder to generate the desired output. 
%
\textbf{(a)} Prior knowledge includes prototypes of digits, which are used to pre-train and build the decoder's generative module, and Sudoku's rules, which help DRNet reason about the overlapping digits.
\textbf{(b)} Reasoning modules batch data points involved in the same constraints (cells in rows, columns, blocks of a Sudoku) together, enforce that the structure of the latent space satisfies prior knowledge, and dynamically adjust the weights of constraints based on their satisfiability.
\textbf{(c)} The overall objective combines responses from the generative decoder (thinking fast) and the reasoning modules (thinking slow). 
}
\label{fig:DRN}
\vspace{-10pt}
\end{figure}

We propose \textbf{Deep Reasoning Networks (DRNets)},
an end-to-end framework that
combines deep learning with logical and constraint reasoning for solving complex tasks that require both \textit{System 1} and \textit{System 2} 
style thinking, 
typically in an unsupervised or weakly-supervised setting.
We illustrate the power of DRNets for disentangling two overlapping hand-written 
 Sudokus 
 (\textbf{Multi-MNIST-Sudoku})
(see Fig.\ref{fig:sudoku_examples}) and for solving a substantially more complex task in scientific discovery that concerns inferring crystal structures of materials from X-ray diffraction data, which we refer to as \textbf{Crystal-Structure-Phase-Mapping}.
%
%
%
%
Both tasks  
require probabilistic reasoning to interpret noisy and uncertain data, while satisfying a set of rules:  
Sudoku rules
and thermodynamic rules. 
For example, de-mixing hand written digits is challenging, but it becomes more feasible when we  reason about the prior knowledge concerning the two overlapping Sudokus. 
Crystal structure phase mapping is yet substantially more complex.
In fact, crystal structure phase mapping easily becomes too complex for experts to solve 
and is a major bottleneck in high-throughput materials discovery.
DRNets are  motivated and inspired by  problems from scientific discovery,  such as  crystal structure phase mapping.

\textbf{Our contributions:} \textbf{(1)} We introduce \textbf{Deep Reasoning 
Networks (DRNets)}, an end-to-end unsupervised framework that combines deep learning with logical and constraint reasoning. 
 DRNets perform end-to-end deep reasoning by encoding a latent space of the input data that captures the structure and prior knowledge constraints within and among data points (Fig.\ref{fig:DRN}). 
The latent space is used by a generative decoder to generate the desired output, 
consistent with the input data and prior knowledge.
DRNets optimize an objective function capturing the overall problem objective as well as prior knowledge in the form of weighted constraints, using \textbf{(2) Constraint-Aware Stochastic Gradient Descent}.
DRNets batch data points involved in the same 
constraint component together and dynamically adjust the constraints' weights as a function of their satisfiability during the optimization
phase. 
\textbf{(3)} We propose a \textbf{group of entropy-based  continuous relaxations that use probabilistic modelling to encode general discrete constraints including sparsity, cardinality, so-called All-Different constraints, and SAT constraints}.
%
De facto, these examples illustrate how to develop “gadgets” to encode a variety of 
combinatorial constraints and prior knowledge in DRNets. 
\textbf{(4)} We show 
\textbf{how to encode Multi-MNIST-Sudoku, standard 9-by-9 Sudoku, SAT, and
  Crystal-Structure-Phase-Mapping as DRNets}, by
properly defining the structure of the latent space, additional
reasoning modules to model the problem constraints (prior knowledge),
and the components of the objective function. \textbf{(5)} We provide
\textbf{ detailed experimental results} demonstrating the potential of
DRNets.  In particular, we show how \textbf{(5.1)} DRNets significantly
  outperformed the state of the art and human experts on
  \textbf{Crystal-Structure-Phase-Mapping instances}, recovering more precise,
  interpretable, and physically meaningful crystal structure pattern decompositions.  \textbf{(5.2)}  On
 \textbf{Multi-MNIST-Sudoku instances}, DRNets perfectly recovered the
  digits in the mixed Sudokus with 100\% digit accuracy and
  outperformed the supervised state-of-the-art MNIST de-mixing models,
including CapsuleNet \citep{sabour2017dynamic} and ResNet
\citep{he2016deep}.
  \textbf{(5.3)} DRNets also solve standard \textbf{combinatorial problems},  such as \textbf{9-by-9 Sudoku puzzles} and \textbf{3-SAT} \citep{mitchell1992hard}, which require hidden structure reasoning, outperforming the supervised deep-learning state of the art. %

While we illustrate the potential of DRNets applied to different variants of Sudoku, 3-SAT problems, and Crystal-Structure-Phase-Mapping, DRNets are general and can be adapted and expanded to many other applications.  Future research entails developing the corresponding “gadgets” for incorporating other types of constraints, prior knowledge, and objective functions, for other applications.
 

%% file: related.tex
\section{Related Work}
Exploiting problem structure and reasoning about prior knowledge in machine learning tasks has been of increasing interest to facilitate learning, enhance generalization, and improve interpretability \citep{taskar2004max,ganchev2010posterior,ermon2015pattern,hu2016harnessing}.
Bayesian machine learning \citep{nasrabadi2007pattern} imposes prior beliefs by regularizing the posterior with prior distributions.
\citet{ganchev2010posterior} proposed \textit{posterior regularization} (PR), which encodes the soft constraints via a variational distribution.
\citet{hu2016harnessing, hu2016deep} introduced the PR framework into deep learning for solving natural language processing tasks.
In computer vision, symmetry and bone-length constraints were introduced for human pose estimation \citep{zhou2017weaklysupervised,zhou2016deep}, and linear constraints were imposed for image segmentation \citep{pathak2015constrained}.
In structured prediction, \citet{chen2018end} imposed a multivariate Gaussian distribution to capture the correlation among multiple entities, and \citet{lee2017enforcing} incorporate constraints at the inference stage via fine-tuning.
In reinforcement learning, \citet{anthony2017thinking, silver2016mastering, silver2018general} outsource the reasoning process (\textit{System 2}) to external Monte Carlo tree search.
In representation learning, k-Sparse autoencoder \citep{makhzani2013k} proposed a k-sparse encoding of the original data.
A PCA-like autoencoder \citep{ladjal2019pcalike} uses a covariance loss term to encourage the dimensions of the latent space to be statistically independent.
Deep generative models \citep{goodfellow2014generative,kingma2013auto,oord2016pixel, larochelle2011neural,hu2017unifying} intrinsically impose a prior distribution into the latent space to reason about the original data distribution, which implicitly exploits the underlying structure.
InfoGan \citep{chen2016infogan} uses mutual information loss to compress most information into an interpretable low-dimensional encoding.
\citet{mirza2014conditional, hu2017toward} use labeled data to control the sample attributes and disentangle the latent space.
\citet{hu2018deep} introduced \textit{posterior regularization} into deep generative models to learn structured knowledge from labeled data that improves the quality of generated samples.

Leveraging machine learning to solve combinatorial optimization problems  has also received much attention (see e.g., \citet{bengio2018machine} for a recent survey). For examples: \citet{bello2016neural} and \citet{bengio2018machine} explored reinforcement learning and Pointer Networks for the \textit{traveling salesman problem}.
\citet{li2018combinatorial} use graph convolutional networks to guide the local search for solving graph-related NP-complete problems.
\citet{selsam2018learning, amizadeh2019pdp}
proposed
NeuroSAT and PDP 
to tackle SAT problems with specialized neural networks and one-bit supervision.
\citet{wilder2018melding} proposed to 
use continuous relaxation of discrete problems to 
backpropagate the gradients,
to upstream machine learning models.

While exploiting problem structure and prior knowledge in deep neural networks has received much attention, previous works primarily focus on supervised settings for data-rich domains, typically using large amounts of labeled data, which reduces the importance of explicitly reasoning about prior knowledge given the data's strong signal.
Furthermore, with few exceptions, the constraints proposed in previous models are often independent, soft, i.e., violating them only leads to a worse solution, and are mainly used as regularization terms.
In contrast, the problems that DRNets aim to solve often involve many constraints that are hard and correlated: satisfying one constraint while neglecting others can potentially make them unsatisfiable, and violating any of them directly results in an invalid solution (e.g., a wrong Sudoku digit could make the whole puzzle unsolvable).
Therefore, a tactical architecture and a smart reasoning module are needed to tackle such challenges. 
To the best of our knowledge, DRNets are the first end-to-end unsupervised framework that combines deep learning with logical and constraint reasoning for solving complex tasks.

%% file: model.tex
\section{Deep Reasoning Networks}
\begin{figure}[t]
  \centering
  \includegraphics[width=13.8cm]{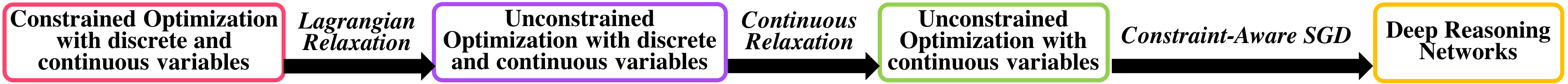}
  \caption{The reduction flow of Deep Reasoning Networks.}
    \label{fig:DRNet-flow}
  \vspace{-15pt}
\end{figure}
DRNets (see Fig.\ref{fig:DRN}) are inspired by human thinking \citep{shivhare2016cognitive}: we abstract patterns to higher-level descriptions and combine them with prior-knowledge to fill-in the gaps. Consider the Multi-MNIST-Sudoku example (Fig.1): we first guess the digits in each cell based on the patterns;  we re-adjust our initial beliefs and re-image the overlapping patterns by reasoning about Sudoku rules and comparing them to the original ones, potentially involving several iterations.{\\
Formally, \textbf{DRNets formulate unsupervised learning as constrained optimization, incorporating abstractions and reasoning about structure and prior knowledge:}
}
\small
\begin{gather}
    \min\limits_{\theta}\; \frac{1}{N}\sum^N_{i=1}{\mathcal{L}(G(\phi_\theta(\mathbf{x}_i)), \mathbf{x}_i)} \quad
    \mbox{ s.t. } \phi_\theta(\mathbf{x}_i) \in \Omega^{\mbox{local}} \mbox{ and }
    (\phi_\theta(\mathbf{x}_1), ..., \phi_\theta(\mathbf{x}_N) ) \in \Omega^{\mbox{global}}
    \label{eqn:constrainted}
\end{gather}
\normalsize
In this formulation,
$\mathbf{x}_i \in R^n $ is the $i$-th $n$-dimensional input data point, $\phi_\theta(\cdot)$ is the function of the encoder in DRNets parameterized by $\theta$, $G(\cdot)$ denotes the generative decoder, $\mathcal{L}(\cdot,\cdot)$ is the loss function (e.g., evaluating the reconstruction of patterns),  $\Omega^{\mbox{local}}$ and $\Omega^{\mbox{global}}$ are the constrained spaces w.r.t.\ 
a single input data point 
and several input data points, respectively.
%
$G(\cdot)$ is 
in general a fixed pre-trained or  parametric model.
For example, in Multi-MNIST-Sudoku, $G(\cdot)$ is a pre-trained conditional GAN \citep{mirza2014conditional} using hand-written digits, and for Crystal-Structure-Phase-Mapping, $G(\cdot)$ is a Gaussian Mixture model.
Note that constraints can involve several (potentially all) data points: e.g., in Sudoku, all digits should form a valid Sudoku and in crystal-structure-phase-mapping, all data points in a composition graph should form a valid phase diagram.
Thus, we specify local and global constraints in DRNets -- local constraints only involve a single input data point whereas global constraints involve several input data points, and they are
optimized using different strategies.

Solving the constrained optimization problem  (\ref{eqn:constrainted}) directly is extremely challenging since the objective function in general involves deep neural networks, which are highly non-linear and non-convex, and  prior knowledge often even involves combinatorial constraints (Fig.\ref{fig:DRNet-flow}).
Therefore, we use Lagrangian relaxation to approximate 
equation (\ref{eqn:constrainted})
with an unconstrained optimization problem, i.e., 
\small
\begin{gather}
    \min\limits_{\theta}\; \frac{1}{N}
    \sum^N_{i=1}{\mathcal{L}(G(\phi_\theta(\mathbf{x}_i)), \mathbf{x}_i)} + \lambda^l\psi^{l}(\phi_\theta(\mathbf{x}_i))
    + \sum^{N_g}_{j=1}\lambda^g_j\psi^g_j(\{\phi_\theta(\mathbf{x}_k)|k\in S_j\})
    \label{eqn:object}
\end{gather}
\normalsize
$N$ is the number of input data points, 
$N_g$ denotes the number of global constraints, $S_j$ denotes the set of indices w.r.t.\ the data points involved in the $j$-th global constraint, and $\psi^l, \psi^g_j$ denote the penalty functions for local constraints and global constraints, respectively, along with their corresponding penalty weights 
$\lambda^l$ and $\lambda^g_j$.
In the following, we propose two mechanisms to tackle the above unconstrained optimization task (Fig.\ref{fig:DRNet-flow}).

\begin{figure}[h]
    \centering
    \includegraphics[width=13.8cm]{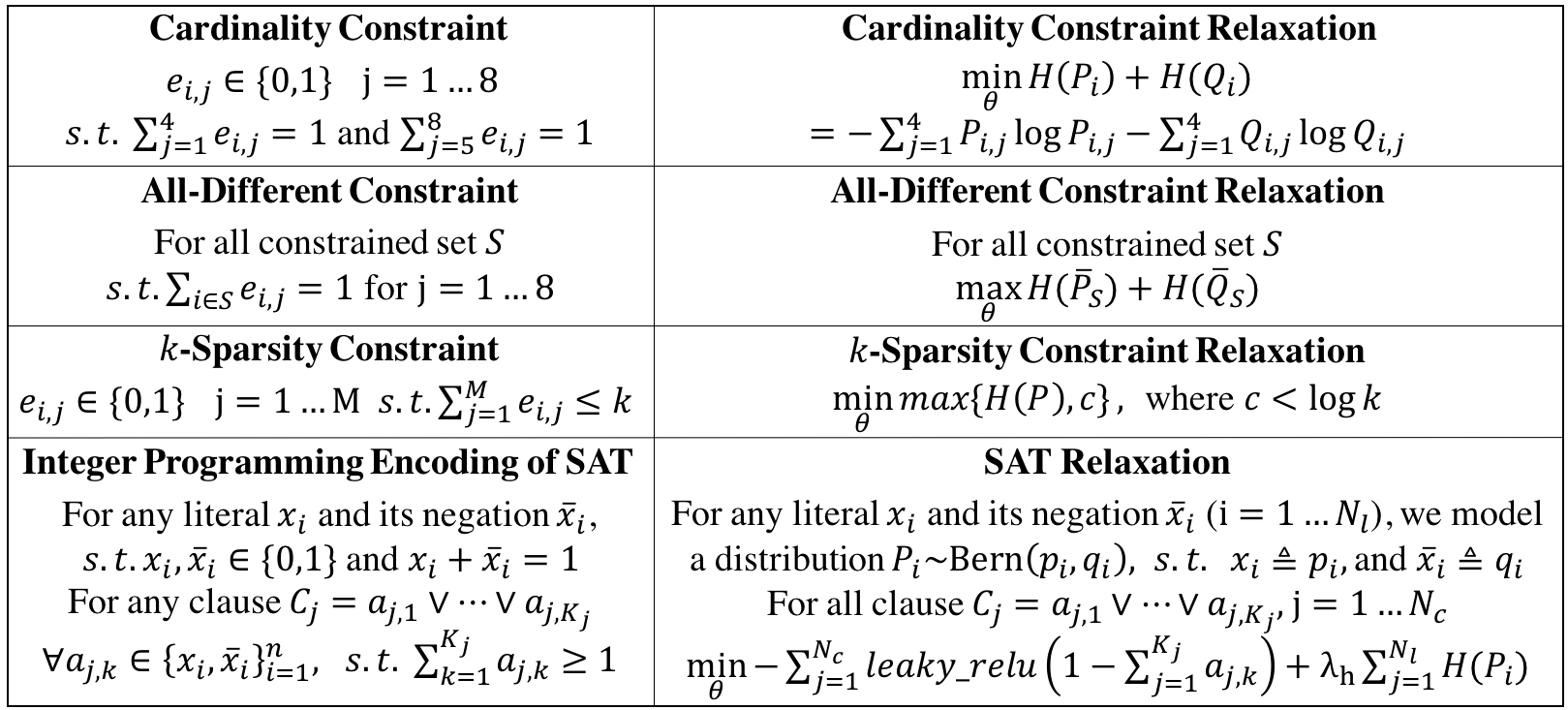}
    \caption{
    Examples of continuous relaxations: $e_{i,j}, N_c, N_l,K_j,\lambda_h$, $P_i$ denote binary variables, the number of clauses, 
    the number of literals, the number of 
    literals in the $j$-th clause, the weights of entropy terms, and the Bernoulli distribution for the $i$-th literal.  "leaky\_relu" is the leaky ReLU.
    }
    \label{fig:relaxation}
\end{figure}

\textbf{Continuous Relaxation:} Prior knowledge often involves combinatorial constraints with discrete variables that are difficult to optimize in an end-to-end manner using gradient-based methods.
Therefore, we need to design proper continuous relaxations for discrete constraints to make the overall objective function differentiable. 
We propose a group of entropy-based 
continuous relaxations to encode general discrete constraints such as sparsity, cardinality, All-Different constraints, and SAT constraints (see Fig.\ref{fig:relaxation}).
Moreover, our framework can be easily expanded to encode other constraints.
We construct continuous relaxations based on probabilistic modelling of discrete variables, where we model a probability distribution over all possible values for each discrete variable. 
For example, in Multi-MNIST-Sudoku, a way of encoding the possible two digits in the cell indicated by data point $x_i$ (one from $\{1...4\}$ and the other from $\{5...8\}$), 
is to use 8 binary variables $e_{i,j}\in\{0, 1\},$
while requiring $\sum^4_{j=1}e_{i,j} = 1$ and $\sum^8_{j=5}e_{i,j} = 1$.
In DRNets, we model probability distribution $P_i$ and $Q_i$ over digits 1 to 4 and 5 to 8 respectively:  $P_{i,j} ,\scriptstyle{j=1...4}$ and $Q_{i,j}, \scriptstyle{j=1...4}$  denote the probability of digit $j$ and the probability of digit $j+4$, respectively.
We approximate the cardinality constraint of $e_{i,j}$ by minimizing the entropy of $P_i$ and $Q_i$, which encourages $P_i$ and $Q_i$ to collapse to one value.\\
Another combinatorial constraint 
in Multi-MNIST-Sudoku is the All-Different constraint, where all the cells in a \textit{constrained set} $S$, i.e., each row, column, and any of four 2x2 boxes involving the corner cells, must be filled with non-repeating digits.
For a probabilistic relaxation of the All-Different constraint, we analogously define the entropy of the averaged digit distribution for all cells in a constrained set $S$, i.e., $H(\bar{P}_S):$
\small
\begin{equation}
    H(\bar{P}_S) =-\sum^4_{j=1}\bar{P}_{S,j}\log{\bar{P}_{S,j}}=
    -\sum^4_{j=1}\bigg(\frac{1}{|S|}\sum_{i\in S}P_{i,j}\bigg)\log\bigg({\frac{1}{|S|}\sum_{i\in S}P_{i,j}}\bigg)
\end{equation}
\normalsize
In this equation, a larger value implies that the digits in the cells of $S$ distribute more uniformly.
Thus, we can analogously approximate All-Different constraints by maximizing $H(\bar{P}_S)$ and $H(\bar{Q}_S)$.
One can see, by minimizing all $H(P_i)$ and $H(Q_i)$ to 0 as well as maximizing all $H(\bar{P}_S)$ and $H(\bar{Q}_S)$ to $\log|S|$, we find a valid solution for the two 4x4 Sudoku puzzles, where all $P_{i,j}$ are either 0 or 1. 
Furthermore, we can easily generalize those two relaxations  for 9x9 Sudoku puzzles.

We also propose to relax the $k$-sparsity constraints, which for example in Crystal-Structure-Phase-Mapping state the maximum number $k$ of pure phases in an XRD-pattern, by minimizing the entropy of the phase distribution $P$ below a threshold $c < \log k$.
We choose the threshold $c < \log k$ because the entropy of a discrete distribution $P$ with at most $k$ positive values cannot exceed $\log k$.

Finally, we approximate the SAT constraints by relaxing their integer programming encoding, where we minimize the entropy of literals to enforce their collapse to either 0 or 1, while maximizing the sum of literals in each clause to encourage one of them to be 1 (true). 
Moreover, we use "leaky\_relu" \citep{xu2015empirical} to discourage increasing the sum in each clause when its larger than 1. 

  

\begin{algorithm}[ht]
\caption{Constraint-aware stochastic gradient descent optimization of deep reasoning networks.
}
\label{alg:training} 
\begin{algorithmic}[1]
\REQUIRE \textbf{(i)} Data points $\{x_i\}_{i=1}^N$. \textbf{(ii)}  Constraint graph. \textbf{(iii)} Penalty functions $\psi^l(\cdot)$ and $\psi_j^g(\cdot)$ for the local and the global constraints.
\textbf{(iv)} Pre-trained or parametric generative decoder $G(\cdot)$.
\STATE Initialize the penalty weights $\lambda^l, \lambda_j^g$ and thresholds for all constraints.
\FOR{number of  optimization iterations}
\STATE Batch data points $\{\mathbf{x}_1,...,\mathbf{x}_m\}$ from the sampled (maximal) connected components.
\STATE Collect the global penalty functions $\{\psi^g_j(\cdot)\}^M_{j=1}$ concerning those data points.
\STATE Compute the latent space $\{\phi_\theta(\mathbf{x}_1), ...,\phi_\theta(\mathbf{x}_m)\}$ from the encoder.
\STATE Adjust the penalty weights $\lambda_l, \lambda_j^g$ and thresholds accordingly.
\STATE minimize $\frac{1}{m}\big(\sum^m_{i=1}{\mathcal{L}(G(\phi_\theta(\mathbf{x}_i)), \mathbf{x}_i)} + \lambda_l\psi^{l}(\phi_\theta(\mathbf{x}_i))\big) +
    \sum^{M}_{j=1}\lambda^g_j\psi^g_j(\{\phi_\theta(\mathbf{x}_k)|k\in S_j\})$
    using any standard gradient-based optimization method and update the parameters $\theta$. 
\ENDFOR
\end{algorithmic}
\end{algorithm}
\textbf{Constraint-Aware Stochastic Gradient Descent:} We propose constraint-aware SGD to tackle the global penalty functions $\psi^g_j(\{\phi_\theta(\mathbf{x}_k)|k\in S_j\})$, which involve several (potentially all) data points.
We define a \textit{constraint graph}, an undirected graph in which each data point forms a vertex and two data points are linked if they are in the same global constraint.
Constraint-aware SGD batches data points from the randomly sampled (maximal) connected components in the \textit{constraint graph}, and optimizes the objective function w.r.t.\ the subset of global constraints concerning those data points and the associated local constraints. 
For example, in Multi-MNIST-Sudoku, each overlapping Sudoku forms a maximal connected component, we 
batch the data points from several randomly sampled overlapping Sudokus and optimize the All-Different constraints (global) as well as the cardinality constraints (local) within them.
However, in Crystal-Structure-Phase-Mapping, the maximal connected component becomes too large to batch together, due to the constraints (\textit{phase field connectivity} and \textit{Gibbs-alloying rule}) concerning all data points in the composition graph. 
Thus, we instead only batch a subset (still a connected component) of the maximal connected component -- e.g., a path in the composition graph, and optimize the objective function that only concerns constraints within the subset (along the path).
By iteratively solving sampled local structures of the "large" maximal component, we cost-efficiently approximate the entire global constraint.
Moreover, for optimizing the overall objective, constraint-aware SGD dynamically adjusts the thresholds and the weights of constraints according to their satisfiability, which can involve non-differentiable functions.

For efficiency, DRNets solve all instances together using constraint-aware SGD (see Algorithm \ref{alg:training}).

%% file: experiments.tex
\section{Experiments}
We illustrate the power of DRNets on two complex tasks and two standard combinatorial problems -- disentangling two overlapping hand-written 
Sudokus (\textbf{Multi-MNIST-Sudoku}), inferring crystal structures of materials from X-ray diffraction data (\textbf{Crystal-Structure-Phase-Mapping}), solving \textbf{9x9 Sudoku Puzzles,} and \textbf{3-SAT problems}. 
We use 3-layer-fully-connected networks as our encoders for all tasks, but we use different generative decoders for different tasks.
Moreover, since DRNets are an unsupervised framework, we can apply the \textit{restart} \citep{gomes1998boosting} mechanism, i.e., we can re-run DRNets  for 
unsolved instances.

\textbf{Multi-MNIST-Sudoku:} 
We generated 160,000 input data points which correspond to 32x32 images of overlapping digits coming from the \textit{test set} of MNIST \citep{lecun1998gradient} and every 16 data points form a 4-by-4 overlapping Sudokus.
Note, our task is more challenging than CapsuleNet's  \citep{sabour2017dynamic}, in which they offset the digits by 4 pixels, while we fully overlap them, explaining CapsuleNet's different performance.
For Multi-MNIST-Sudoku, the DRNet batches every 16 data points together to enforce the All-Different constraints among the cells of each Sudoku.
We use a conditional GAN \citep{mirza2014conditional} as our generative decoder (denoted as $G(\cdot)$), which is trained using the digits in the \textit{training set} of MNIST. 
For each cell $\mathbf{x}_i$, the decoder encodes a latent space, which consists of two parts:  
The first part includes two distribution $P_i$ and $Q_i$ (see Fig.\ref{fig:MMS-2}) concerning the possible digits in the cell, and the second part is the latent encodings $z_{i,1},...,z_{i,8}$ of each possible digit conditioned on the overlapping digits, which is used by the generative decoder to generate the corresponding digits $G(z_{i,j})$. 
We obtain our estimation of the two digits in the cell by computing the expected digits over $P_i$ and $Q_i$, i.e., $\sum^4_{j=1}{P_{i,j}G(z_{i,j})}$ and  $\sum^4_{j=1}{Q_{i,j}G(z_{i,j+4})}$, and reconstruct the original input mixture (see Fig.\ref{fig:MMS-2}). 
As we described before, we impose the continuous relaxation of cardinality constraints and All-Different constraints to reason about the Sudoku structure among cells of the overlapping Sudokus.
To demonstrate the power of reasoning, we compared our unsupervised DRNets with supervised start-of-the-art MNIST de-mixing models -- CapsuleNet \citep{sabour2017dynamic} and ResNet \citep{he2016deep}, and a variant of DRNets that removes the reasoning modules ("DRNets w/o Reasoning"). 
We evaluate both the percentage of digits that are correctly de-mixed (digit accuracy) and the percentage of overlapping Sudokus that have all digits correctly de-mixed (Sudoku accuracy).
Empowered by reasoning, DRNets significantly outperformed CapsuleNet, ResNet, and DRNets without reasoning, perfectly recovered all digits with the \textit{restart} mechanism (see Table \ref{table:MMS}), and
additionally reconstructed the mixture with high-quality (see Fig.\ref{fig:sudoku_examples}).

\begin{minipage}[b]{\textwidth}
\begin{minipage}[b]{0.60\textwidth}
\centering
\includegraphics[height=3.3cm]{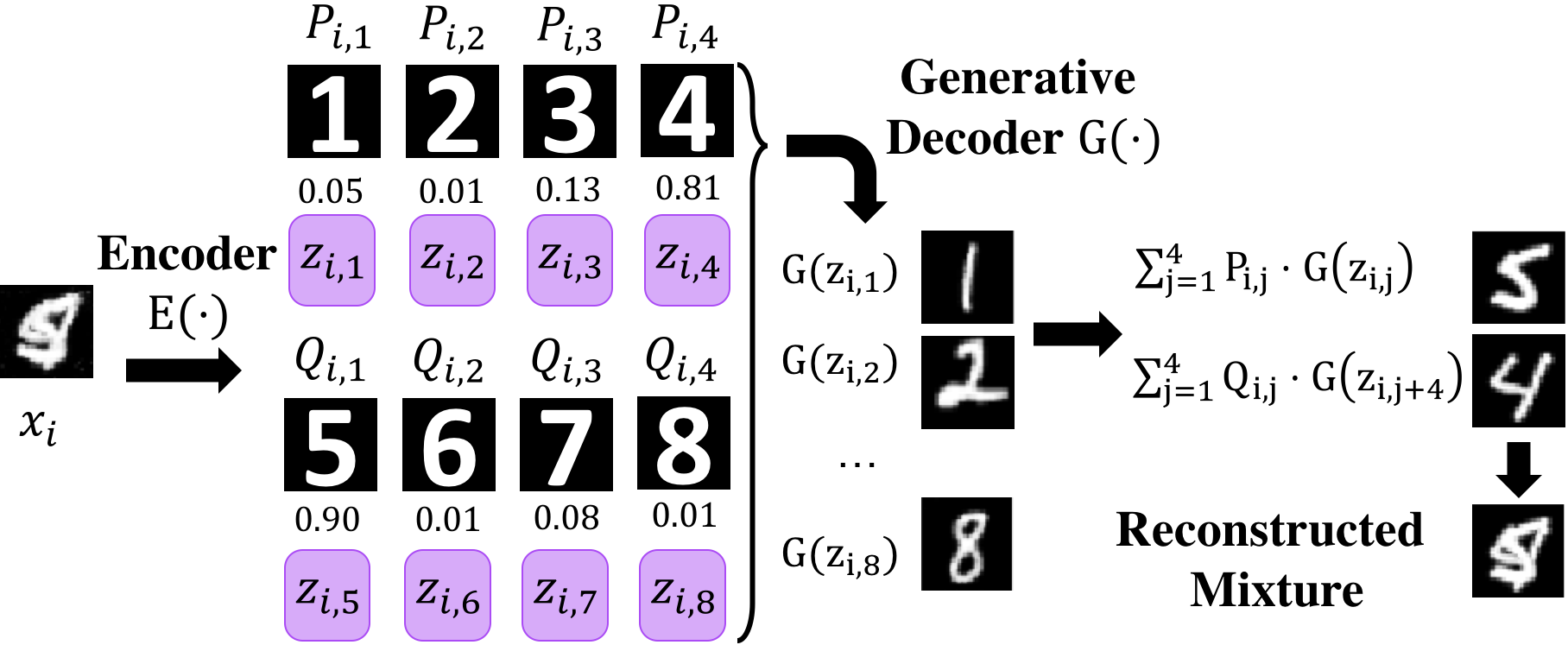}
\captionof{figure}{The latent space of the DRNet for Multi-MNIST-Sudoku.}
\label{fig:MMS-2}
\end{minipage}
\hfill
\begin{minipage}[b]{0.35\textwidth}
\centering
\footnotesize
\centering
\setlength\tabcolsep{1pt}
\resizebox{\columnwidth}{!}{
\begin{tabular}{|c|c|c|c|}
\hline
\multicolumn{4}{|c|}{\textbf{Multi-MNIST-Sudoku (10,000 instances)}}  \\
\hline
\textbf{Accuracy (\%)} & \textbf{Digit} & \textbf{Sudoku}  & \textbf{Time}\\ 
\hline
\raggedright{\makecell{DRNets + \\ Restart}} & \textbf{100.00} & \textbf{100.00} & 2hours \\
\hline
DRNets & \textbf{99.99} & \textbf{99.92} & 2hours \\
\hline
\raggedright{\makecell{DRNets  w/o \\ Reasoning}} & 90.43 & 20.06 & 2hours \\
\hline
CapsuleNet &  88.46 & 2.01 & 2min + 7hrs    \\
\hline
ResNet-110 & 91.44 & 76.40 & 5min + 1day \\
\hline
\end{tabular}
}
\captionof{table}{
Accuracy comparison. 
We show  "test time + training time" for supervised baselines and  "solving time" for unsupervised DRNets.
}
\label{table:MMS}
\end{minipage}
\end{minipage}

\begin{figure}[ht]
    \centering
    \includegraphics[width=13.8cm]{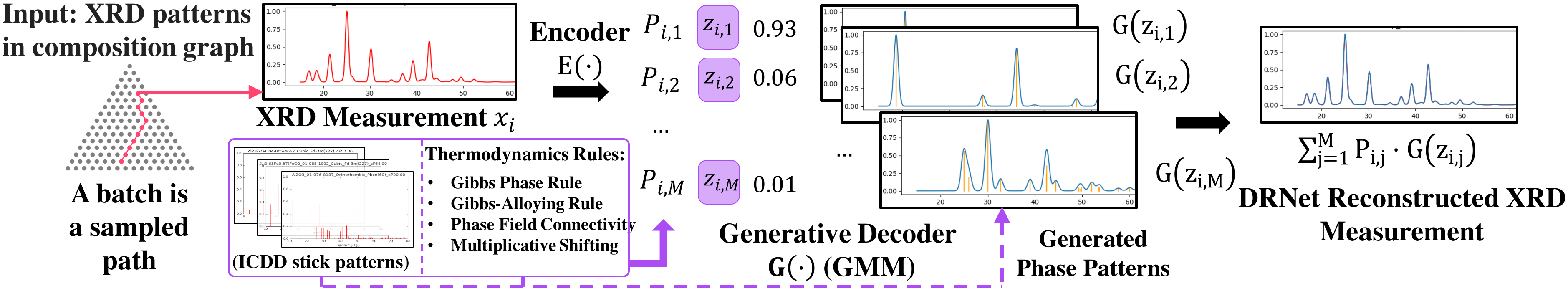}
    \caption{The latent space of the DRNet for Crystal-Structure-Phase-Mapping. $M$ denotes the number of possible phases. (For Al-Li-Fe, $M=159$; For Bi-Cu-V, $M=100$.)}
    \label{fig:XRD-recon}
    \vspace{-5pt}
\end{figure}


\textbf{Crystal-Structure-Phase-Mapping} concerns inferring crystal structures from a set of X-ray diffraction measurements (XRDs) of a given chemical system, given  a variety of thermodynamic  constraints.
Crystal structure phase mapping is a very challenging task:
Each X-ray measurement may involve several mixed crystal structures; each chemical system includes hundreds of possible crystal structures; for each crystal structure pattern, we only have a theoretical (idealized) model of pure crystal phases;
the rules of thermodynamics are also complex; 
and the crystal patterns are difficult for human experts to interpret, much more complex than identifying digits.
In fact, the current state of the art of crystal structure phase mapping  is a major bottleneck in high-throughput materials discovery.


Herein, we illustrate DRNet for crystal structure phase mapping for two chemical systems: (1) a ternary \textbf{Al-Li-Fe} oxide system \citep{le2014challenges}, 
which is theoretically based, synthetically generated, with ground truth solutions, and (2) a ternary \textbf{Bi-Cu-V} oxide system, which is a more challenging real system obtained from chemical experiments and is more noisy and uncertain.
For each system, 
each input data point is the XRD of a mixture of crystal structures. Additionally, the input includes the \textit{composition graph} specifying elemental
compositions and the \textit{constraint graph} of the data points.
We also collected a library of possible crystal structures from the International Centre for Diffraction Data (ICDD) database.
Each crystal structure (also named \textit{phase}) is given as a list of diffraction peak location-amplitude pairs, (referred to as \textit{stick pattern}), representing the ideal phase patterns measured in a perfect condition (see Fig.\ref{fig:XRD-recon}).
To model more realistic conditions, DRNets simulate the real phase patterns from \textit{stick patterns} using Gaussian mixture models, where the relative peak locations and mixture coefficients are given by the stick locations and amplitudes.
Moreover, the peak width, peak location shift, and peak amplitude variance are parameterized by the latent encoding $z_{i,j}$ and used by the generative decoder to generate the corresponding possible phase patterns in the reconstructed XRD measurement.

We compared DRNets with the  state-of-the-art model (IAFD) \citep{bai2017relaxation}, 
which uses non-negative matrix factorization (NMF), interacting with external mixed-integer programming modules to enforce prior knowledge. 
For the Al-Li-Fe oxide system, though IAFD enforced thermodynamic rules, the gap between the external optimizer and NMF resulted in a solution that is far from the ground truth (see Fig.\ref{fig:phase-diagram}). In contrast, DRNet almost exactly recovered the ground truth solution by seamlessly integrating  pattern recognition, reasoning, and prior knowledge, including the novelty of explicitly incorporating the stick pattern information. 
DRNets solved the Bi-Cu-V oxide system, producing valid crystal structures and significantly outperforming IAFD w.r.t. reconstruction error. 
In addition, none of the IAFD phases matched the ICDD stick patterns, indicating that the Bi-Cu-V oxide system is beyond IAFD's capabilities.
Materials science experts thoroughly checked DRNet's Bi-Cu-V-O solution, and approved it. 
They were particularly excited about the results given that the phase map for the \textbf{Bi-Cu-V-O system was previously unknown}, despite their considerable efforts.

\begin{minipage}[b]{\textwidth}
\begin{minipage}[b]{0.65\textwidth}
\centering
\includegraphics[height=2.8cm]{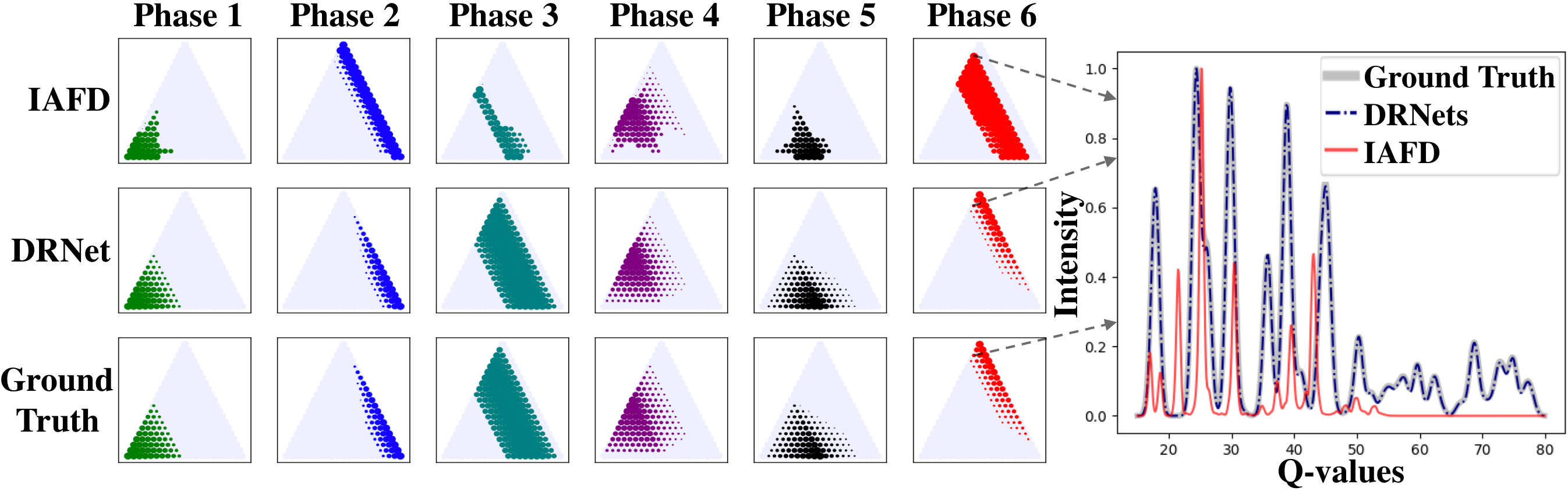}
\captionof{figure}{Comparison of phase concentration 
in Al-Li-Fe oxide system estimated by IAFD and DRNets.
Each dot represents an XRD measurement whose size is proportional to the estimated phase concentration. 
DRNet's phase patterns closely match the ground truth in contrast to IAFD's (see e.g., phase 6, right panel).
}
\label{fig:phase-diagram}
\end{minipage}
\hfill
\begin{minipage}[b]{0.32\textwidth}
\footnotesize
\begin{tabular}{|c|c|c|}
\hline
\textbf{Al-Li-Fe} & L1 loss & L2 loss\\ 
\hline
DRNets & \textbf{0.039} & \textbf{< 0.001}\\
\hline
IAFD & 8.549 & 1.125  \\
\hline
\textbf{Bi-Cu-V} & L1 loss & L2 loss\\ 
\hline
DRNets & \textbf{3.993} & \textbf{0.196} \\
\hline
IAFD & 10.580 & 0.992  \\
\hline
\end{tabular}
\captionof{table}{
DRNets outperform IAFD both on L1 and L2. Num.\ of XRDs: Al-Li-Fe 231; Bi-Cu-V 353; Num.\ of stick patterns: Al-Li-Fe 159; Bi-Cu-V 100.}
\label{table:XRD-recon}
\hfill
\end{minipage}
\hfill
\end{minipage}

\textbf{Combinatorial Problems:} As a proof of concept 
of 
how DRNets can encode standard combinatorial problems, we solve 
9-by-9 Sudoku puzzles and Boolean satisfiability problems (SAT), using a  3-layer-fully-connected network as our encoder and the reasoning modules.
We generated 10,000 9-by-9 Sudoku puzzles with 24 to 32 clues \citep{minimumsudoku} (e.g., see Fig.\ref{fig:sudoku_examples}) and 10,000 satisfiable random 3-SAT instances with the hardest ratio (\#clauses/\#literals = 4.3) \citep{mitchell1992hard}.
%
%
%
%
%
%
We compared DRNets with the supervised deep learning state of the art: Recurrent Relational Networks (RRNets)  \citet{palm2017recurrent}, NeuroSAT \cite{selsam2018learning} (SAT)
and PDP \citep{amizadeh2019pdp} (SAT).  
DRNets, \textit{without supervision}, outperformed all supervised deep learning models (see Table \ref{table:SAT}).

\begin{table}[ht]
    \centering
    \setlength\tabcolsep{2pt}
    \resizebox{\columnwidth}{!}{
    \begin{tabular}{|c|c|c|c|c|c|c|}
    \hline
    \textbf{Instances (10,000)} & DRNets & DRNets + Restart & NeuralSAT & PDP & RRNets\\ 
    \hline
    3-SAT n=30 m=129 & 81.0\% (4min) & \textbf{99.0\%} (33min) & 45.5\% (2min+1hr) & 78.9\% (5min+2hr)  & NA  \\
    \hline
    3-SAT n=50 m=215 & 63.3\% (7min) & \textbf{94.0\%} (47min) & 26.1\% (3min+1hr) & 62.2\% (8min+2hr) & NA\\
    \hline
    3-SAT n=100 m=430 & 34.7\% (17min) & \textbf{77.9\%}  (2hr) & 4.7\% (5min+1hr) & 31.4\% (2hr+2hr) & NA \\
    \hline
    9x9 Sudoku & 99.5\% (1hr)& \textbf{99.8\%} (1hr) & NA & NA & 99.6\% (1min + 1day)\\
    \hline
    \end{tabular}
    }
    \caption{Percentage of instances solved for 3-SAT (hardest ratio $m/n=4.3$) and standard 9x9 Sudoku (24 to 32 known cells).
    We show the "test time + training time" for supervised baselines and the "solving time" for our unsupervised DRNets.
    $n, m$ denote the number of literals and clauses.
    NA, not applicable. DRNets, without supervision, outperform the supervised state of the art.}
    \label{table:SAT}
\end{table}
\vspace{-10pt}

Finally, we stress that our main goal is to tackle problems that combine deep learning and reasoning, such as de-mixing Sudokus or crystal structure phase mapping, as opposed to competing with pure, highly specialized state-of-the-art SAT solvers that can solve larger 3-SAT instances than the ones reported here.
Nevertheless, our results show that DRNets can encode a broad range of combinatorial constraints and prior knowledge and effectively combine deep learning with reasoning.

See supplementary materials for 
further details on DRNets' model and experimental results.

%% file: conclusion.tex
\section{Conclusions and future work}
We propose DRNets, a powerful end-to-end framework that combines deep learning with logical and constraint reasoning for solving complex tasks.  
DRNets outperform the state of the art for de-mixing MNIST Sudokus and crystal-structure phase mapping, solving previously unsolved systems substantially beyond the reach of other methods and materials science experts' capabilities. 
DRNets also outperform the deep-learning state of the art for solving standard Sudokus and 3-SAT.
While we illustrate the potential of DRNets with unsupervised settings, it is straightforward to impose supervision into DRNets.
Future research includes exploring DRNets for incorporating other types of constraints, prior knowledge, and objective functions, for other applications.